\documentclass{IEEEcsmag}
\usepackage[colorlinks,urlcolor=blue,linkcolor=blue,citecolor=blue]{hyperref}
\usepackage[utf8]{inputenc}
\usepackage{upmath}
\usepackage{amsfonts}
\usepackage[
backend=biber,
style=ieee,
]{biblatex}

\DeclareUnicodeCharacter{0301}{\'{}}

\begin{document}

\sptitle{}

\title{Communicating Uncertainty in Machine Learning Explanations: A Visualization Analytics Approach for Predictive Process Monitoring}

\author{Nijat Mehdiyev}
\affil{German Research Center for Artificial Intelligence (DFKI) and Saarland University, Saarbrücken, 66123, Germany}

\author{Maxim Majlatow}
\affil{German Research Center for Artificial Intelligence (DFKI) and Saarland University, Saarbrücken, 66123, Germany}

\author{Peter Fettke}
\affil{German Research Center for Artificial Intelligence (DFKI) and Saarland University, Saarbrücken, 66123, Germany}
\markboth{}{}

\begin{abstract}
\looseness-1
As data-driven intelligent systems advance, the need for reliable and transparent decision-making mechanisms has become increasingly important. Therefore, it is essential to integrate uncertainty quantification and model explainability approaches to foster trustworthy business and operational process analytics. This study explores how model uncertainty can be effectively communicated in global and local post-hoc explanation approaches, such as Partial Dependence Plots (PDP) and Individual Conditional Expectation (ICE) plots. In addition, this study examines appropriate visualization analytics approaches to facilitate such methodological integration. By combining these two research directions, decision-makers can not only justify the plausibility of explanation-driven actionable insights but also validate their reliability. Finally, the study includes expert interviews to assess the suitability of the proposed approach and designed interface for a real-world predictive process monitoring problem in the manufacturing domain. 
\end{abstract}      
\maketitle

\chapteri{T}he widespread use of machine learning (ML) models in real-world, high-stakes decision-making requires establishing reliability and understandability, which promote trust among relevant stakeholders \cite{slack2021reliable}. In this regard, numerous explainable artificial intelligence (XAI) techniques have recently been proposed, including post-hoc explanation approaches, to provide global and local explanations of the model behavior \cite{arrieta2020explainable}. These approaches aim to enhance the transparency and interpretability of the models, enabling stakeholders to comprehend their decision-making processes better.

Uncertainty quantification (UQ) is another emerging ML research field focusing on estimating and communicating the uncertainty associated with models' predictions. It can be regarded as a complementary form of transparency that can boost the explainability of solutions for decision tasks, which may not be sufficient on their own \cite{bhatt2021uncertainty}. When properly calibrated and effectively communicated, uncertainty can improve stakeholders' ability to determine when to trust model predictions and thereby enhance decision automation or augmentation \cite{tomsett2020rapid}. Recent research indicates that providing information on model confidence, in conjunction with appropriate communication strategies, can also offer a layer of transparency and trust for domain experts \cite{zhang2020effect}.

Selecting and deploying appropriate methods to generate explanations or quantify uncertainties is critical in ML model inspection. However, the efficiency of this process might be hampered if the insights are not presented to stakeholders clearly and concisely. Therefore, to accomplish optimal communication of the model's outcomes, it is essential to build interactive interfaces that are customized to the mental models of the intended stakeholders while minimizing their cognitive load. To this end, interactive information visualization has emerged as a promising area of research for enhancing ML models' interpretability, trustworthiness, and reliability by providing users with relevant visual representations \cite{choo2018visual}. Upon closer examination of the relevant literature, it becomes evident that a considerable amount of research has been conducted on either visual analytics for uncertainty communication \cite{hullman2018pursuit}, or model interpretability \cite{alicioglu2022survey}, but in isolation from each other. Despite the importance of both research directions in facilitating more informed decision-making, there appears to be a gap in integrated approaches that combine information visualization for uncertainty communication with model interpretability for a more holistic understanding.

With the gap mentioned above in mind, we present a novel methodology that aims to integrate model uncertainties into both local and global post-hoc explanations. Our approach is deployed in a visual analytics interface that enables to verify the appropriateness and usability of the generated explanations. To ensure that the solution meets the needs of its intended users, a comprehensive requirements analysis for the design process and evaluation has been conducted within a consortium research project. To summarize our main contributions, we have made the following accomplishments:

\begin{itemize}
    \item We explore different presentation forms for communicating model uncertainty for individual predictions, including visualization and textual and tabular descriptions.
    \item We propose and formalize methods to integrate model uncertainty into global post-hoc explanation approaches, specifically PDP, and their local version, ICE plots.
    \item We design visualization analytics instruments to support the implementation of our proposed approach and conduct a comprehensive evaluation with domain experts in a real-world use case involving a predictive process monitoring problem.
\end{itemize}

\section{BACKGROUND AND RELATED WORK}
In numerous practical application scenarios, black-box ML algorithms are essential to reach a level of accuracy that conventional, intrinsically interpretable  ML approaches cannot. Nevertheless, such opaque approaches frequently fail to explain their working mechanism, making it difficult for analysts to verify their veracity \cite{arrieta2020explainable}. Providing explanations is an effective method for promoting acceptance of the predictions provided by intelligent systems. As a result, XAI has arisen as a fruitful area of research to enhance the collaboration between AI-based systems and human users by making the underlying non-transparent algorithms understandable \cite{guidotti2018survey}. The notion of explainability is intricate and multifaceted, requiring consideration of various factors within the decision-making environment. These factors include the analytical context, user attributes, explanation objectives, and a range of socio-cognitive and process-specific aspects. Therefore, when designing intelligent methods and interfaces, it is crucial to take into account these factors to ensure adequate explainability \cite{mehdiyev2021explainable}. 

Different taxonomies are available for XAI techniques, one of the main categories being post-hoc explanation techniques. These techniques provide explanations for AI model predictions and can be grouped into two categories: local and global explanations. Local techniques (SHapley Additive exPlanations (SHAP), Local Interpretable Model-Agnostic Explanations (LIME), ICE) explain individual predictions, while global methods (PDP, SHAP Summary Plots, Permutation-based Feature Importance) explain the overall behavior of an AI model \cite{islam2022systematic}. Post-hoc explanation techniques are often preferred by domain experts for justification and verification purposes due to their comprehensible nature. However, recent studies have revealed that most of these post-hoc explanation techniques exhibit inconsistency and instability and may fail to provide adequate information regarding their reliability, highlighting the need for integrating model uncertainty estimation \cite{slack2021reliable}. Nonetheless, to date, only a limited number of studies undertook the endeavor to model uncertainty in post-hoc explanations, as we do in this study \cite{slack2021reliable,moosbauer2021explaining,cafri2016understanding}. Moreover, there has been a significant lack of attention towards developing user interfaces in this intersection that incorporate visualization analytics approaches.

The ML lifecycle encompasses multiple stages, from data collection to model training, each of which introduces inherent uncertainties. As a result, the predictions generated by AI models are subject to various types of uncertainty, such as errors in data collection, model complexity, and algorithmic limitations. Two types of uncertainty that are commonly distinguished in the context of AI models are aleatoric uncertainty, and epistemic uncertainty \cite{gawlikowski2021survey}. Aleatoric uncertainty is related to inherent data variability or the observed phenomena' underlying stochastic nature. In contrast, epistemic uncertainty arises from incomplete or insufficient knowledge about the modeled system or the model's limitations. While aleatoric uncertainty depicts uncertainty that cannot be reduced, usually pertaining to the underlying data, epistemic uncertainty is reducible and either due to incomplete data or a characteristic of the fitted model. Methods used to quantify the uncertainty that capture both categories can be divided into two main categories: Bayesian approaches and Frequentist approaches  \cite{bhatt2021uncertainty}. An overview of techniques from both families can be found in this study \cite{gawlikowski2021survey}.

Integrating UQ in the context of explainability is a step towards a holistic, trustworthy Artificial Intelligence (AI), especially regarding the user's trust and acceptance of a model's decisions.   

\section{RESEARCH METHODOLOGY}
This section describes the approach to constructing an uncertainty-aware XAI solution with corresponding interfaces. The design science research (DSR) approach is adopted to provide a systematic and rigorous process for conducting applied research \cite{peffers2007design}. This methodology is particularly suitable for information systems research and involves six steps: problem identification and motivation, defining objectives of a solution, design and development, demonstration, evaluation, and communication.

\textbf{Problem Identification and Motivation:} To secure both contextual applicability and methodological precision, our article draws upon inputs from the application domain for contextual relevance and the existing scientific knowledge base for methodological rigor. To ensure relevance, a consortium research method is used to engage practitioners in identifying open issues and defining objectives. Furthermore, an extensive literature analysis was conducted to secure the rigor, and findings were refined through iterative discussions with practitioners.

The primary challenge identified in this study is to develop an approach that can overcome the non-transparent nature of black-box ML techniques. This challenge is particularly relevant in high-stakes decision-making problems. To tackle this issue, the design of such systems should incorporate post-hoc explanation techniques that enable the users to ratify the validity of model decisions. Moreover, to ensure that solutions are not only explainable but also reliable and trustworthy, it is essential to effectively communicate model uncertainty in the explanations generated by the system

\textbf{Objective of a Solution:} The next step involves defining the objectives of the solution. The goals have two dimensions. First, a methodological approach is required for communicating model uncertainty and incorporating this information into post-hoc explanations. This would increase the reliability and trust of underlying algorithms. Second, interfaces with relevant visualization techniques should be devised to effectively communicate the outcomes to the system users.

\textbf{Design and Development:} Our proposed artifact comprises a deep feedforward neural network for generating predictions, Monte-Carlo (MC) dropout to estimate model uncertainty, and ICE plots and PDP approach to generate local and global explanations, respectively. The study's novelty lies in incorporating uncertainty information into these visualization-based post-hoc explanation approaches. With such integration, decision-makers can not only justify the plausibility of explanation-driven actionable insights but also validate their reliability by examining the model confidence.

\textbf{Demonstration:} The applicability of the proposed artifact has been examined for a predictive process monitoring problem. Predictive process monitoring is a branch of process mining that combines advanced computational intelligence methods with process modeling approaches \cite{fettke2020modelling}. The objective is to enable continuous business process improvement by extracting predictive, data-driven, process-specific insights from the event logs generated by a process-aware information system (PAIS) \cite{van2012bprocess}. Event logs are an essential enabler for evidence-based process analysis by providing necessary details about process execution.

More specifically, we address a real-world use case scenario in the manufacturing domain. The examined problem pertains to cycle time prediction, with a focus on predicting the duration of individual manufacturing activities required to fulfill orders. The data is obtained from the Manufacturing Execution Systems (MES) of the consortium partner. Prior to implementing our proposed artifact, we conducted a rigorous feature engineering process using process-specific data that had been enriched with customer order data.

\textbf{Evaluation:} The evaluation phase involves conducting semi-structured interviews with two domain experts who provide critical and constructive feedback on the system design, usability, and suggestions for improvements. The evaluation results provide valuable insights to refine the system and improve its usability.

\textbf{Communication:} Finally, the communication phase involves sharing the findings and approach details through scientific publications and industrial events to a broad audience of researchers and practitioners from different backgrounds. 

Through the use of DSR, we can ensure a methodologically sound and systematic process for conducting applied research that is relevant to the application domain. Ultimately, the XAI solution developed using this approach, with a focus on communicating uncertainty, is expected to enhance the trustworthiness of AI systems. 

\section{UNCERTAINTY ESTIMATION IN POST-HOC EXPLANATIONS}
In this section, we provide an overview of the mathematical foundations that underpin our proposed novel uncertainty-aware XAI approach. 
\subsection{Uncertainty Quantification with Monte Carlo Dropout}
MC dropout technique, proposed by \cite{gal_dropout}, is a state-of-the-art method for quantifying uncertainty in deep learning models. Conventionally, the dropout technique is incorporated during the training phase of a neural network to prevent overfitting by randomly dropping or ignoring neurons. Enabling dropout regularization during the deployment phase can be seen as a Bayesian approximation to the probabilistic deep Gaussian process. By performing $T \in \mathbb{N}^+$ stochastic forward-passes through the network, the model's predictive variance can be calculated, which can serve as a measure of uncertainty.

We realize MC dropout by performing $T \geq 50$ stochastic forward passes through our deep feedforward neural network. This approach allows the model's predictions to be mapped to the corresponding variance. In particular, uncertainty profiles can be created for the training data by first sorting the variances in ascending order, then calculating the variance thresholds. For example, thresholds for the 25th and 75th percentile can be calculated, resulting in three uncertainty profiles. The utilization of percentile-based estimations provides an initial foundation for the categorization of model confidence profiles. Ultimately, the domain experts hold the responsibility of determining the final profiles, either by refining the initial data-driven estimations or by defining their own ranges or categories.                      
\subsection{Uncertainty-Aware Partial Dependence Plots}
The PDP approach, as proposed by \cite{friedman_pdp}, is a global explanation method used to depict a fitted model's learned relationship between predictor variables and a target variable: Given a set of predictor variables, create a new data set by replacing the values of these predictors with those of one specific observation and let the fitted model score this new data set. Then, by averaging the resulting scores and repeating this process for the rest of the observed feature values, the mean predictions can be plotted against the observed feature values, resulting in the corresponding PDP. 

Let $\mathbf{x} = \left(x_{1}, ..., x_{p}\right)$ be a set of predictor variables and $y$ a response variable, let 
$D=\left\{\left(\mathbf{x_i}, y_i\right)\right\}_{i=1}^N$ be the data set of associated $\left(\mathbf{x}, y\right)$-values, with N denoting total amount of observations. The data set is being divided into three data sets $D=D_{train} \cup  D_{val} \cup D_{test}$, with $D_{train}$ being used for training, $D_{val}$ for hyperparameter optimization, and $D_{test}$ for evaluation, with $N_{train}$, $N_{val}$, $N_{test}$ be the respective amount of observations for each subset. Let $\widehat{F}(\mathbf{x})$ be the function provided by the machine learning model, mapping $\mathbf{x}$ to $y$ and fitted on $D_{train}$. For $S\subset\{1, ..., p\}$ let $\mathbf{x}_{S} \subset \mathbf{x}$ be a subset of predictor variables for which a PDP is going to be generated, with $C\subset\{1, ..., p\}\setminus S$ being the complement of $S$ and $\mathbf{x}_{C} \subset \mathbf{x} \setminus \mathbf{x}_{S}$ being the complement of the selected predictor variables $\mathbf{x}_{S}$. The partial dependence of $\widehat{F}\left(\mathbf{x}\right) = \widehat{F}\left(\mathbf{x}_{S},\mathbf{x}_{C}\right)$ on the selected predictors ${x}_{S}$ is defined as
\begin{equation}
\overline{F}_S\left(\mathbf{x}_S\right)=E_{\mathbf{x}_{C}}[\widehat{F}(\mathbf{x})]=\int \widehat{F}\left(\mathbf{x}_S, \mathbf{x}_{C}\right) p_{C}\left(\mathbf{x}_{C}\right) d \mathbf{x}_{C}
\end{equation}
with $p_{C}(\mathbf{x}_{C})=\int p(\mathbf{x}) d \mathbf{x}_S$ being the marginal probability density of $\mathbf{x}_{C}$, and $p(\mathbf{x})$ being the joint density of $\mathbf{x}$. Estimating $p_{C}(\mathbf{x}_{C})$ from the training data allows (1) to be expressed as
\begin{equation}
\overline{F}_S\left(\mathbf{x}_S\right)=\frac{1}{N_{train}} \sum_{i=1}^{N_{train}} \widehat{F}(\mathbf{x}_S, \mathbf{x}_{C,i}).
\end{equation}

The practical implementation of PDP that incorporates UQ can be realized as follows:

Let $\mathbf{x}_S$ be the predictors of interest with unique values $\{\mathbf{x}_{S,1}\, ..., \mathbf{x}_{S,j}\}$, $j$ denoting the amount of unique observed values from the training data and let $T$ be the amount of stochastic forward passes:
\begin{enumerate}
    \item For $k \in \{1, 2, ..., j\}$:
    \begin{enumerate}
        \item Create a copy $\widehat{D}_{train}$ of $D_{train}$ and replace each original value of $\mathbf{x}_S$ within $\widehat{D}_{train}$ with the specific value $\mathbf{x}_{S,k}$.
        \item Compute the average model prediction $\overline{F}_S(\mathbf{x}_{S,k})$ based on the vector of predicted values for $\widehat{D}_{train}$.
        \item Compute the vector $v_k$ of variances through $T$ stochastic forward passes using the UQ approach.
        \item Compute the distribution of uncertainty profile members from $v_k$, saving the identification of the most represented group in $color_k$.     
    \end{enumerate}
    \item Plot the pairs $\{\mathbf{x}_{S,k}, \overline{F}_S(\mathbf{x}_{S,k})\}$ to receive the PDP.
    \item Color the pairs $\{\mathbf{x}_{S,k}, \overline{F}_S(\mathbf{x}_{S,k})\}$ in accordance to the corresponding $color_k$-values.
    \item For each pair $\{\mathbf{x}_{S,k}, \overline{F}_S(\mathbf{x}_{S,k})\}$, plot the distribution of variances $v_k$ as a density plot or histogram and make the chart accessible for the user when selecting a point.
    \item For each pair $\{\mathbf{x}_{S,k}, \overline{F}_S(\mathbf{x}_{S,k})\}$, plot the distribution of uncertainty profile members from $v_k$ using a doughnut chart or histogram and make the chart accessible for the user when selecting a point.
\end{enumerate}

Steps c) and d) are necessary for the incorporation of the proposed UQ approach, expanding on the approach of \cite{friedman_pdp}, whereas 3), 4), and 5) are the proposed visualizations for human-in-the-loop analysis (see Fig. 2 and Fig. 3). This PDP depicts the average effect of a set of selected features on the prediction of the fitted model, however, as presented by \cite{goldstein_ice}, the averaging of predictions for a specific feature value can still conceal local relationships between the feature of interest and the target. To uncover such relationships, we propose the incorporation of UQ into ICE plots.   

\subsection{Uncertainty-Aware Individual Conditional Expectation Plots}
An ICE plot, as proposed by \cite{goldstein_ice}, is a global explanation method that focuses on a single feature and depicts changes in a fitted model's predictions when said feature is being marginally altered. In particular, \cite{goldstein_ice} proposes the following approach: Given a selected set of predictor variables with $k$ unique values and one observation from the underlying data set, expand this observation to a new, artificial data set by copying it $k$-times and replacing the values of the selected predictors by iterating over their observed unique values. Then let the fitted model score this new data set, resulting in a vector of prediction scores that can be mapped to the corresponding unique predictor values, allowing the relationship to be plotted. If only one predictor is selected, this relationship can be visualized, for example, via a line plot. By repeating this process for the rest of the observations from the underlying data set, one line for each observation of the data set can be added to the plot. Averaging those lines results in the corresponding PDP for the selected predictors. Consequently, using an ICE plot for a single observation represents a local explanation method.

The incorporation of UQ into ICE plots is proposed as a local explanations method, focusing on a single observation from the underlying data set as well as a single predictor of interest, and can be realized as follows: 
Let ${x}_S$ be the predictor of interest with unique values $\{x_{S,1}, ..., x_{S,j}\}$, $j$ denoting the amount of unique observed values from the training data and let $(x^{(i)}_S, \mathbf{x}^{(i)}_C), i \in \{1, ..., N\}$ be a single observation selected from the underlying data set:
\begin{enumerate}
    \item For $k \in \{1, 2, ..., j\}$:
    \begin{enumerate}
        \item Create a copy of the selected instance and replace each original value of $x^{(i)}_S$ with the specific value $x_{S,k}$.
        \item Compute the model prediction $\widehat{F}(x_{S,k}, \mathbf{x}^{(i)}_C)$.
        \item Compute the variance $v^{(i)}_k$ for $(x_{S,k}, \mathbf{x}^{(i)}_C)$ through $T$ stochastic forward passes using the UQ approach.
        \item Compute the membership of $v^{(i)}_k$ to the uncertainty profile, saving the identifier of the group in $color^{(i)}_k$. 
    \end{enumerate}
    \item Plot the pairs $\{x_{S,k}, \widehat{F}(x_{S,k}, \mathbf{x}^{(i)}_C)\}$ to receive the ICE plot.
    \item Color the pairs $\{x_{S,k}, \widehat{F}(x_{S,k}, \mathbf{x}^{(i)}_C)\}$ in accordance to the corresponding $color^{(i)}_k$-values.
    \item Respectively plot the distribution of feature values and predictions as stacked histograms, grouped and colored by the affiliation with the corresponding uncertainty profile, displayed at the corresponding axis.
\end{enumerate}                     

\section{INTERFACE OVERVIEW}
The main focus of this section is to introduce the primary interface components of the solution that are specifically designed to effectively communicate uncertainties arising with each model decision, along with corresponding explanations. Apart from a component for a general overview and instance selection ({\bf Figure 1}), the three interface components that make up the solution are "Uncertainty Estimation and Visualization," "Uncertainty Communication in ICE Plots," and "Uncertainty Communication in PDP."

The core ML components for this project are built using the \textit{"keras"} library in R. Other key libraries utilized for data preparation, interface creation, and visualization include \textit{"data.table", "dplyr", "ggplot2", "plotly", "vip", "shinydashboard"} etc.                                             
\subsection{Uncertainty Estimation and Visualization}

The "Uncertainty Estimation and Visualization" component of our proposed solution is designed to inform system users about model uncertainty using various presentation forms. More specifically, two distinct visualization techniques were utilized, a density plot and a box plot, along with textual and tabular descriptions that convey information on the specific model uncertainty for the selected prediction. 

To begin, the first visualization approach presents the distribution of possible model predictions generated using the chosen UQ approach (such as MC dropout) through density plots ({\bf Figure 2}, A). This allows users to visually inspect distribution details and understand the ranges where the model predictions are predominantly located. Alternatively, we can use a box plot to visualize the same information (Figure 2, C), showing the interquartile range within its hinges, a vertical line representing the median, and whiskers extending to the lowest and highest data points within 1.5 times the interquartile range. Both visualization approaches are supplemented with additional information. For instance, prediction intervals are incorporated into the plots, showing the range within which the predictions will fall with a 95 \% probability. A label below each plot includes the confidence interval as a visual aid, depicted using arrows pointing at the red vertical lines. The plots are colored based on the qualitative uncertainty descriptions, with green, yellow, and red representing "low," "medium," and "high" confidence profiles, respectively.

In addition to these visualization approaches, we provide a textual description (Figure 2, B) that includes information about the UQ method, an explanation of prediction intervals in words, and the uncertainty profile that shows the model confidence for the particular prediction. Finally, a table is presented to the user  (Figure 2, D), showing the model prediction, standard deviation, and other relevant information.
\subsection{Uncertainty Communication in Individual Conditional Expectation Plots}

This component of our proposed uncertainty-aware local explanation approach provides an interface that visualizes prediction scores for new synthetic instances and communicates their uncertainty information ({\bf Figure 3}, B for numerical and {\bf Figure 4}, B for categorical variables). This is achieved through the use of color-coded confidence intervals within the plot. In addition, the uncertainty for each synthetic instance is communicated through the various presentation forms described in the previous component. 

Furthermore, we include histograms as complementary charts at the appropriate axis of the ICE plot to depict the distribution of prediction scores and feature values, enabling users to discern the placement of the original value of the examined prediction instance within the broader data set. Through this analysis, users can identify whether the prediction instance's value is within a frequently occurring range of values or represents an atypical outlier.     
\subsection{Uncertainty Communication in Partial Dependence Plots}
To improve the transparency and interpretability of the PDP, we introduce a new component for the PDP interface that provides two types of uncertainty information ({\bf Figure 5}, B). The first type of uncertainty information is presented through a complementary density plot (Figure 5, C), which displays the distribution of predictions and the 95 \% confidence level bands around them. While this information could be directly visualized in the main graph, we found that doing so would make it harder for users to read and interpret the plot. 
The main contribution of our approach is the visualization of model uncertainty for each examined value of the feature of interest within PDP analysis. To achieve this, we use a doughnut chart (Figure 5, D) that displays the distribution of uncertainty profiles, giving users an overview of the model's global reliability for the examined value.      

\section{EVALUATION}       
\subsection{Usage Scenario}

The effectiveness of the designed interface, which includes visualization analytics components, is showcased in this usage scenario. The interface assists process experts (PE) in estimating the cycle time required for a given customer order. The sequence of required production activities is already predetermined based on the order specifications. However, the PE must still determine the duration of each activity, which can be aided by our data-driven approach. As the timely completion of high-priority orders is critical, the PE are responsible for ensuring the validity of the data-driven guidance provided by the system.

\begin{figure*}
\centerline{\includegraphics[width=36pc]{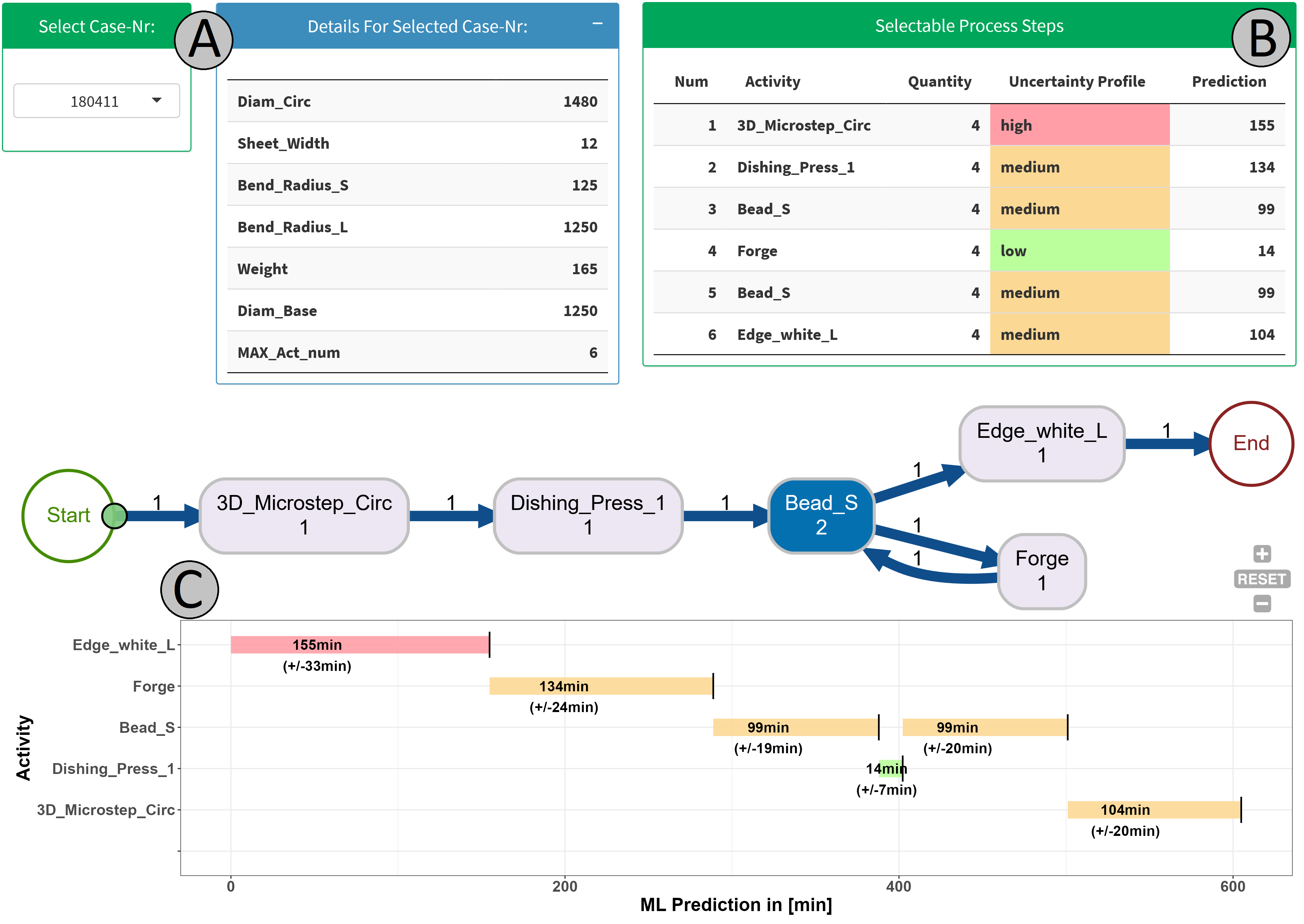}}
\caption{General overview. A: The drop-down menu on the upper left-hand side allows the user to select a production case; corresponding details are displayed on the right-hand side. B: A table view allows further insight into each step of the production case and information concerning activity duration prediction and uncertainty. Process activities are selectable for further analysis. C: An animated process map and a Gantt-Chart depict the planned sequencing and cycle time prediction.}
\end{figure*}

\textbf{General Overview:} The system interaction commences with the PE being directed to a dedicated "General Overview" page exclusively designed for the predictive process monitoring use case. The PE can select the relevant case of interest on this page by using a drop-down menu corresponding to a customer order's production activity sequence (Fig. 1, A). The page also highlights the specifications of the customer order, such as product weight and dimensions, which are also used as inputs to the ML algorithm. Fig. 1, B shows all pertinent information regarding duration prediction for each production step, along with uncertainty profile information. An animated process map and a Gantt-Chart (Fig. 1, C) provide a visual representation of the activity sequence, with the latter featuring the predicted duration for each process step. Each step in the Gantt-Chart is color-coded according to one of three uncertainty categories: "high" is indicated by the color red, "medium" by the color orange, and "low" by the color green.

The production activities in this particular scenario include 3D-cutting ("3D\_Microstep\_Circ"), work at a dishing press ("Dishing\_Press\_1"), beading ("Bead\_S"), shape adjustments in the forge ("Forge"), another beading ("Bead\_S") and refinement of the edges ("Edge\_white\_L"). Our solution categorizes activities 2, 3, 5, and 6 as having a "medium" uncertainty profile, whereas activity 3 has a "low" uncertainty profile. The first activity, 3D-cutting, is an exception and falls under the "high" uncertainty group. It is predicted to take 155 minutes with a standard deviation of 33 minutes, as shown in the Gantt-Chart. The PE decides to investigate this production activity since any disruption at this early stage can potentially cause a cascading effect on the rest of the processes.

\begin{figure*}
\centerline{\includegraphics[width=36pc]{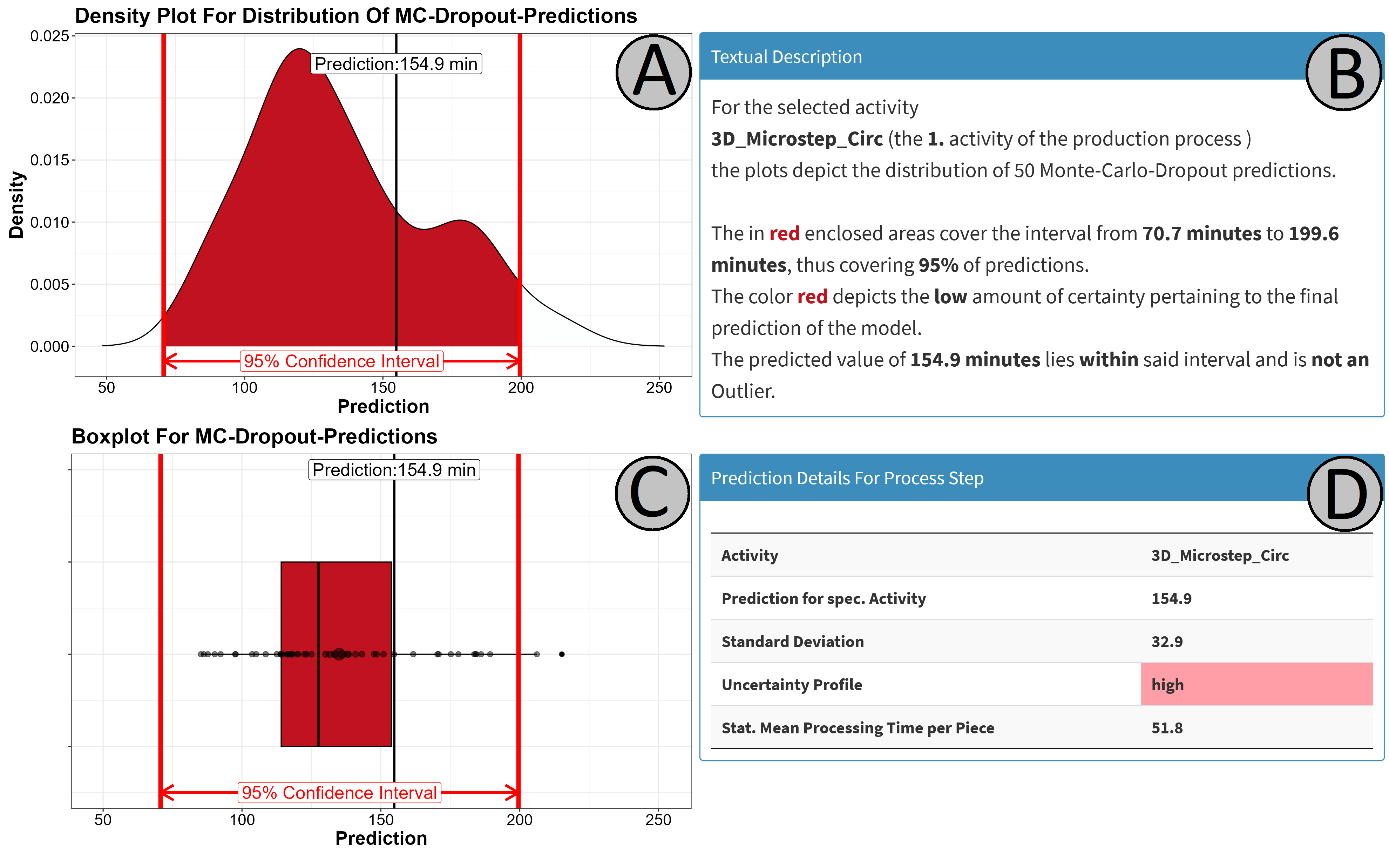}}
\caption{Dashboard interface for uncertainty analysis on an instance-level. A: The density plot depicts the distribution of MC Dropout predictions. B: The same data from A is visualized as a box plot. C: A textual description summarizes important findings from A and B. D: A table displays additional information concerning the examined production activity. The red color coding depicts the "high" level of uncertainty affiliated with the activity duration prediction.}
\end{figure*}

\textbf{Uncertainty Analysis on the Activity Level:} The PE can analyze the uncertainty of individual activities by clicking on the activity of interest, in this case, 3D-cutting. This action displays the distribution of MC dropout predictions as shown in Fig. 2. The model prediction of 154.9 minutes falls into the highlighted confidence interval of 70.7 minutes to 199.6 minutes (Fig. 2, A), which covers 95\% of the values and indicates that the model prediction is not an outlier. However, the upper hinge of the box plot (Fig. 2, C) and the upper limit of the confidence interval suggest that a delay of nearly 45 minutes is not unlikely. A textual description (Fig. 2, B) is also provided to ensure correct interpretation. Finally, Fig. 2, D provides a tabular summary for quick access to duration predictions and uncertainty information for the chosen activity.

\textbf{Uncertainty-aware ICE Plots:} To understand the impact of both numerical and categorical features on model predictions and generate a plan to avoid undesired outcomes, the PE consults the uncertainty-aware ICE plot (Fig. 3, A). By analyzing variables related to the product specifications, such as the bend radius ("Bend\_Radius\_S"), the PE realizes that the duration prediction of 3D-cutting activity around the original value (125 cm) results in "high" uncertainty predictions (Fig. 3, B). However, increasing the value of this feature reduces the predicted activity duration and increases the model's confidence in its predictions, resulting in "medium" uncertainty. This uncertainty-aware ICE plot enables the PE to understand the relationship between the feature of interest and model outcomes and comprehend the model confidence. However, the value of this feature can not be altered for the planning process as it is a fixed, predetermined product specification.

The PE identifies a categorical variable, "Worker," that can be manipulated without affecting the order requirements (Fig. 4, A) and filters out unavailable personnel. In this scenario, the PE notices that the prediction for the allocated worker by the system (anonymized through the identifier "751") falls into the "high" uncertainty category (Fig. 4, B), while other workers would be available with a "medium" uncertainty prediction. Within the area with the lowest prediction scores, the PE chooses a group of three other available workers ("114", "736", and "797") whose predicted duration falls into the "medium" uncertainty group to examine them further. The anticipated durations for these three workers are \textasciitilde103, \textasciitilde118, and \textasciitilde123 minutes for "797", "736," and "114," respectively.

The PE validates the accuracy of the model outputs by cross-referencing them with their domain expertise. The improvement in predictions resulting from the alteration of the selected worker is attributed to the greater experience of the alternative workers in performing the production activity. Since the model's estimated prediction duration and uncertainty align with the PE's expectations, their confidence in its reliability increases.

\begin{figure*}
\centerline{\includegraphics[width=36pc]{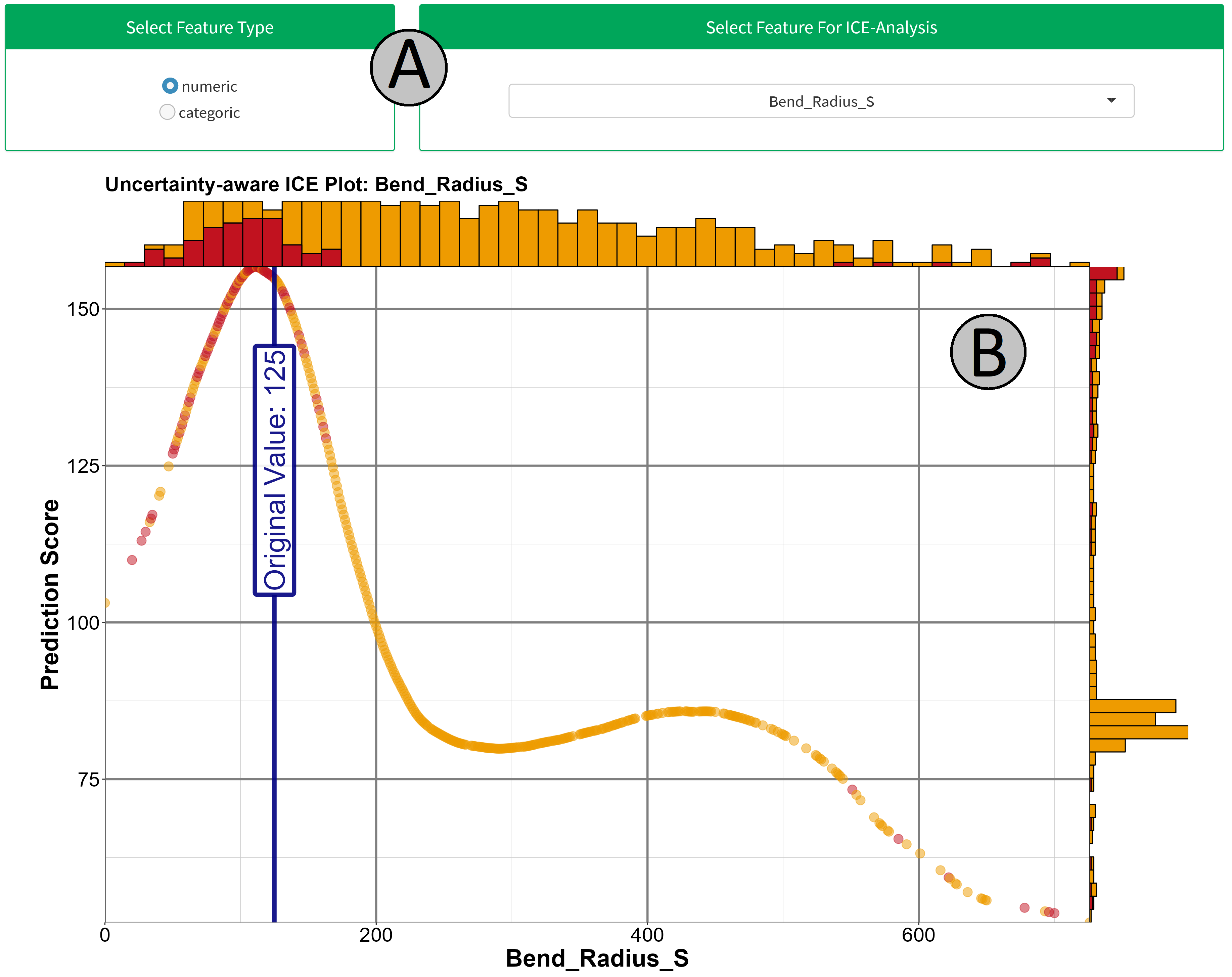}}
\caption{Dashboard interface for uncertainty-aware ICE plots. A: The user selects "numerical" as the variable type on the upper left-hand side, updating the drop-down menu with variables of the chosen type right next to it. B: The proposed uncertainty-aware ICE plot for numerical variables is displayed here.}
\end{figure*}

\begin{figure*}
\centerline{\includegraphics[width=36pc]{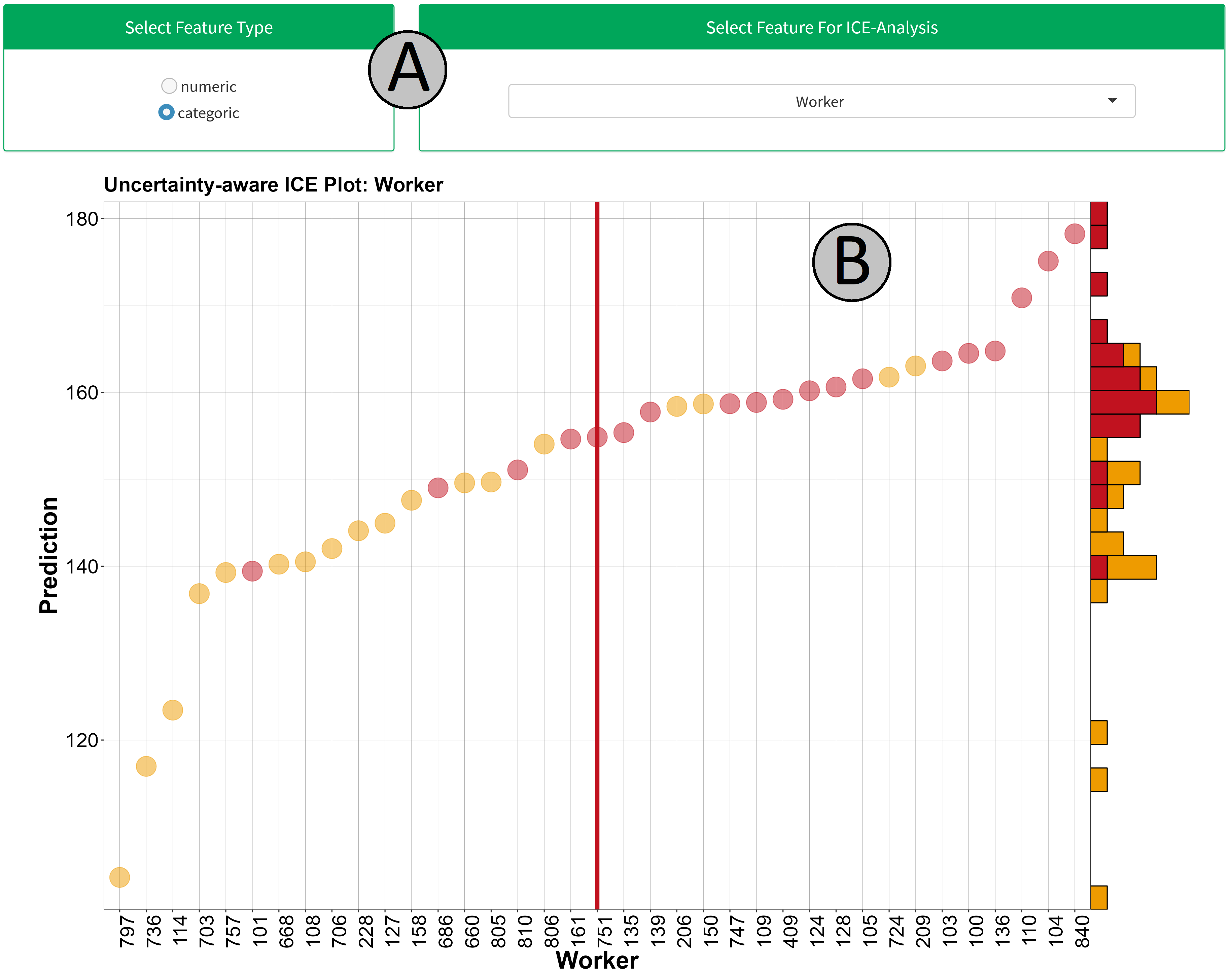}}
\caption{Dashboard interface for uncertainty-aware ICE plots. A: The user selects "categorical" as the variable type on the upper left-hand side, updating the drop-down menu with variables of the chosen type right next to it. B: The proposed uncertainty-aware ICE plot for categorical variables is displayed here, with the distribution of categorical values being omitted since each value only occurs once during plot generation. The red vertical line indicates the original variable value of the analyzed instance.}
\end{figure*}

\textbf{Uncertainty-aware PDP:} The PE switches to the "Global Explanations" tab to ensure that at least one of the selected replacement workers performs well in general, given the high priority of the order. This tab provides the PE with a permutation-based feature importance plot (Fig. 5, A) which helps in understanding the overall impact of certain variables.
Additionally, an uncertainty-aware PDP (Fig. 5, B) is also presented to the PE as a tool for further analysis.

 The PE selects the categorical variable "Worker" and iteratively examines each of the three available workers. By using the distribution of the MC dropout predictions (Fig. 5, C), the PE concludes that worker "797" has an upper boundary of the confidence interval that is approximately 30 minutes lower than the currently allocated worker, with a mean prediction score (82 minutes) that is 22 minutes lower. Furthermore, the doughnut chart (Fig. 5, D) indicates that worker "797" is associated with a greater amount of "low" (33.1\%) and a smaller amount of "high" (16.6\%) uncertainty when compared to the current worker (21.2\% "low," 22.7\% "high" uncertainty). Consequently, the PE has sufficient grounds to modify the production plan by substituting the current worker "751" with worker "797" for the given process activity, reducing the predicted lead time and decreasing model uncertainty.

\begin{figure*}
\centerline{\includegraphics[width=37pc]{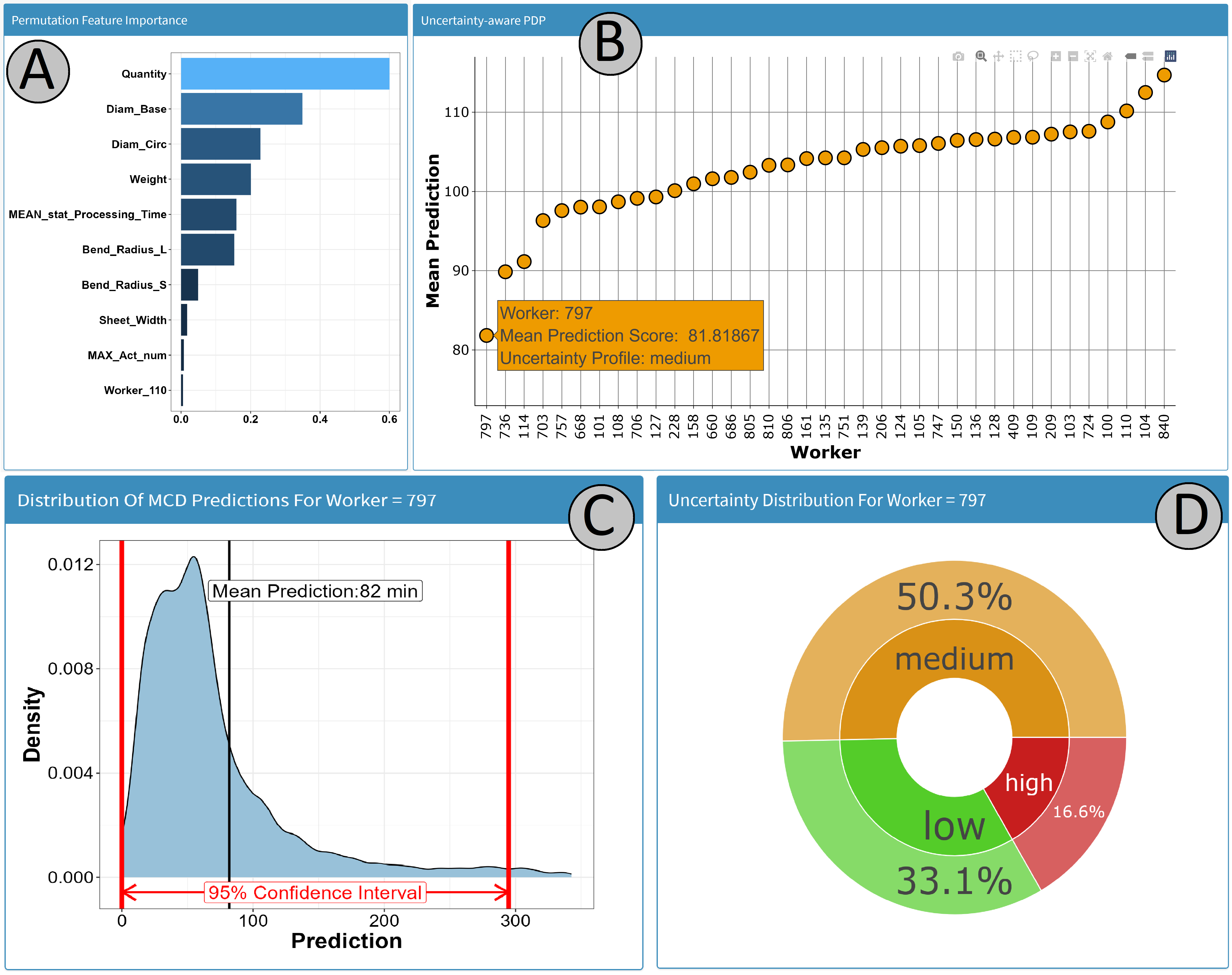}}
\caption{Dashboard interface for uncertainty-aware PDP. A: Permutation-based feature importance is displayed as a supporting tool for orientation and global explainability. B: The proposed uncertainty-aware PDP for categorical variables is displayed here. Hovering over a point displays its mean prediction score and the corresponding dominant uncertainty group. C: Clicking on a point in B updates this density plot, which displays the distribution of predicted values. D: Clicking on a point in B updates this doughnut chart, which displays the distribution of uncertainty group membership from the corresponding predicted values.}
\end{figure*}
\subsection{Expert Interview}
During the evaluation of the proposed uncertainty-aware explanation methods, an interview was conducted with two process experts, including a factory manager and a production manager. Both experts possess extensive knowledge of the underlying data and have a deep understanding of the processes involved, as well as expert knowledge regarding the interrelationships between the features under examination and the target.

During the interview, the process experts selected an exemplary Case-ID and explored the visualizations for the corresponding production process to establish a starting point for further analysis. They quickly orientated themselves within this tab and delved into the analysis of individual production steps. Within the exemplary case, they chose a process step and explored the uncertainty analysis in detail. Following that, uncertainty-aware ICE plots were presented and discussed in a similar manner. The global explanations, which contain the uncertainty-aware PDP, were examined next. These steps were repeated for the other process steps within the exemplary case. Next, they performed the same steps for other randomly chosen cases and were asked to provide feedback on how they would interact with the dashboard if it were deployed for production planning. Finally, they discussed the dashboard in detail, and the expert feedback was documented. The interview lasted for one hour, and each step described above took approximately 15 minutes. The interviewed experts provided valuable feedback on the usability and design of the visualization and explanation techniques employed in the proposed uncertainty-aware explanation methods and proposed suggestions for improvement. 

\textbf{Design and Usability:} In terms of design, the experts expressed positive views toward the proposed uncertainty-aware ICE plots and uncertainty-aware PDP and were able to derive and validate the relationships depicted in these visualizations shortly after their introduction. Furthermore, the integrated color scheme for faster differentiation of uncertainty profiles was internalized quickly and utilized by the experts as they explored real-world usage scenarios in the third interview step.

The density plot was rated as the most accessible and effective visualization method for the distribution of MC dropout predictions, followed by the box plot as a complementary visualization. The additional textual explanations were highlighted as a powerful tool for preventing false interpretations and improving overall user acceptance. The experts found the relationship between the prediction score and the distribution of MC dropout predictions easy to grasp after a brief initial explanation.

\textbf{Suggestions for Improvement:} While the experts were generally satisfied with the proposed explanation methods, they expressed a desire for the ability to compare similar cases. This would enable users to filter the underlying data to construct uncertainty-aware ICE plots and uncertainty-aware PDP restricted based on the filtered data. Incorporating this feature into the proposed dashboard would enhance the user experience and improve the utility of the proposed explanation methods for production planning. 

\section{DISCUSSION AND CONCLUSION}

In this work, we presented an approach for integrating and communicating model uncertainty in the context of XAI, focusing on visualization as a medium for conveying information. In particular, we introduced uncertainty-aware ICE plots as a local and uncertainty-aware PDP as a global explanation method, enhanced with various visual properties and functionalities for deeper uncertainty analysis. We presented the efficacy of this approach in a real-world manufacturing scenario, demonstrating its utility in the hands of a process expert. Additionally, an interview with experts evaluated the effectiveness and usability of the presented methods when integrated into a prototypical dashboard. Below we discuss findings, challenges, and future work.

\textbf{Flexibility and Transferability:} Our study employed a deep neural network and utilized MC dropout to quantify uncertainties in the model's predictions accurately. The use of data-driven methodologies in these steps is not limited to MC dropout and can be interchanged with other methods, such as Extreme Gradient Boosting (XGBoost) for generating predictions or the bootstrap approach for uncertainty quantification. The flexibility of our approach makes it applicable to a wide range of classification or regression challenges that involve tabular data, not just limited to predictive process monitoring scenarios.

\textbf{Combining Local and Global Explanations:} The integration of instance-based and global explanations have been found to increase overall trust and acceptance of AI models. By combining global and local explanations, users can gain a high-level overview of the model and the ability to delve into the details of specific instances. Compared to providing only global or local explanations, this approach is considered more effective in promoting a better understanding and trust in AI models. Our evaluation further underscores the importance of using both explanations for exploring model uncertainty.

\textbf{Scalability:}  The integration of uncertainty UQ and XAI techniques, such as MC Dropout and ICE and PDP, presents significant computational challenges, particularly concerning scalability. For example, the complexity of generating ICE and PDP plots for models with high-dimensional inputs can result in enormous evaluations, which can be prohibitively expensive when MC Dropout generates multiple predictions for each input. In addition, the computational cost of combining these techniques can make it difficult to scale to larger datasets or more complex models. To mitigate these challenges, several approaches have been proposed. One potential solution is to parallelize computations across multiple GPUs, which can lead to significant speedups. This approach is practical for MC Dropout, where the numerous forward passes required for each input can be efficiently distributed across multiple processors. Additionally, binned values can be used for XAI techniques instead of all unique values, which can significantly reduce the number of evaluations required.

\textbf{Future Work:} Based on the recommendations of domain experts and lessons learned during the research and conception phase, several future research directions have been identified. One of these directions pertains to evaluating both UQ and XAI methods. Given that there are currently no standard procedures or mechanisms for assessing the combination of these techniques, this poses a particular challenge. One potential avenue of future research to address this challenge is to extend the dashboard by integrating uncertainty reliability measures such as Prediction Interval Coverage Probability (PICP) and Mean Prediction Interval Width (MPIW). Depending on the results of these measures, different calibration techniques can be employed to improve the reliability of the models. In addition, the dashboard interface can be extended to include model accuracy measures such as Mean Absolute Error (MAE) or Root Mean Squared Error (RMSE) for different uncertainty profiles. This will give users a more comprehensive understanding of the model performance under varying levels of uncertainty. Another interesting area of future research is the design of decision augmentation and automation scenarios using the uncertainty-aware explanations generated. By leveraging these explanations, users can make more informed decisions in high-stakes scenarios. Finally, the combined UQ and XAI methods can be explored to examine the algorithmic fairness of decision-making in high-stakes use cases. Given the potential for bias and discrimination in AI-based decision-making systems, this research direction has significant social and ethical implications. Overall, these future research directions will advance our understanding of the performance and reliability of UQ and XAI methods and inform the development of more robust and equitable decision-making systems.

\section{REFERENCES} 
\printbibliography[heading=none]

\begin{IEEEbiography}{Nijat Mehdiyev}{\,} is a senior researcher at the German Research Center for Artificial Intelligence (DFKI). He holds a bachelor's degree in Business Information Systems from the University of Siegen and a master's degree with honors in Finance and Information Management (FIM) from the Technical University of Munich and the University of Augsburg. Nijat completed his doctoral dissertation on the topic ‘Explainable Artificial Intelligence for Predictive Process Analytics — A Conceptual Framework, Methods and Applications’ at Saarland University.  His research interests include human‐centered explainable artificial intelligence, information systems research, predictive process analytics and data‐driven decision-making. He is actively engaged in national research projects in the Industry 4.0 domain and co‐authored several academic publications. Contact: nijat.mehdiyev@dfki.de.
\end{IEEEbiography}

\begin{IEEEbiography}{Maxim Majlatow}{\,} is a researcher at the German Research Center for Artificial Intelligence (DFKI) since 2022. He received his bachelor's degree in Business Informatics with a focus on machine learning in 2021 from the Saarland University. His primary research interest lies in explainable artificial intelligence, visualization and user-interface development in the context of real-world industrial use cases. Contact: maxim.majlatow@dfki.de.
\end{IEEEbiography}

\begin{IEEEbiography}{Peter Fettke} {\,} is a professor of Business Informatics at Saarland University and principal researcher, research fellow and research group leader at the German Research Center for Artificial Intelligence (DFKI) in Saarbrücken, Germany. In his application‐oriented basic research and basic‐oriented applied research, he and his research group of around 30 people address questions at the interface of the disciplines of business informatics and artificial intelligence (AI), in particular the modeling of computer‐integrated systems, systems mining, process predictions, and robotic process automation. His work is among the most cited articles in leading international journals on business informatics and he is one of the top 5 most cited scientists at DFKI. He is Co‐Editor‐in‐Chief of the journal ‘Enterprise Modelling and Information Systems Architectures’ (EMISAJ). Peter Fettke founded the DFKI‐based ‘Center of Competence for Tax Technology’ and ‘Competence Center for Audit Transformation’. Contact: peter.fettke@dfki.de.
\end{IEEEbiography}

\end{document}